\documentclass[review]{elsarticle}

\usepackage{lineno,hyperref}
\modulolinenumbers[5]

\journal{Journal of \LaTeX\ Templates}

\usepackage{soul}
\soulregister\cite7
\soulregister\ref7
\usepackage{color,xcolor}
\usepackage{multirow}
\usepackage{subfigure}
\usepackage{threeparttable}
\usepackage{graphicx}
\usepackage{makecell}

\begin{document}

\begin{frontmatter}

\title{STCSNN: High energy efficiency spike-train level  spiking neural networks with spatio-temporal conversion}


\author[mymainaddress,mysecondaryaddress]{Changqing Xu\corref{mycorrespondingauthor}}
\cortext[mycorrespondingauthor]{Corresponding author}
\ead{cqxu@xidian.edu.cn}

\author[mysecondaryaddress]{Yi Liu}
\author[mysecondaryaddress]{Yintang Yang}

\address[mymainaddress]{Guangzhou Institute of Technology, Xidian University, Xi’an 710071, China.}
\address[mysecondaryaddress]{School of Microelectronics, Xidian University, Xi’an 710071, China.}

\begin{abstract}
Brain-inspired spiking neuron networks (SNNs) have attracted widespread research interest due to their low power features, high biological plausibility, and strong spatiotemporal information processing capability. 
Although adopting a surrogate gradient (SG) makes the non-differentiability SNN trainable, achieving comparable accuracy for ANNs and keeping low-power features simultaneously is still tricky.
In this paper, we proposed an energy-efficient spike-train level spiking neural network with spatio-temporal conversion, which has low computational cost and high accuracy.
In the STCSNN, spatio-temporal conversion blocks (STCBs) are proposed to keep the low power features of SNNs and improve accuracy.
However, STCSNN cannot adopt backpropagation algorithms directly due to the non-differentiability nature of spike trains.
We proposed a suitable learning rule for STCSNNs by deducing the equivalent gradient of STCB.
We evaluate the proposed STCSNN on static and neuromorphic datasets, including Fashion-Mnist, Cifar10, Cifar100, TinyImageNet, and DVS-Cifar10.
The experiment results show that our proposed STCSNN outperforms the state-of-the-art accuracy on nearly all datasets, using fewer time steps and being highly energy-efficient.
\end{abstract}


\end{frontmatter}


\section{Introduction}

Spiking neural networks (SNNs) have gained increasing attention recently due to their biologically-inspired computational principles\cite{xu2022direct,yang2022robust}, spatiotemporal information processing capability, high robustness\cite{yang2023sibols,yang2023spike}, and potential for achieving high energy efficiency\cite{xu2020boosting,yang2023snib} in neuromorphic computing.
However, the training of SNNs is still a significant challenge\cite{meng2022training,yang2023effective} since information in SNNs is transmitted and processed through non-differentiable spike trains.
To tackle this problem, the surrogate gradient (SG) method\cite{xu2022ultra, xu2023ultra, zhang2019spike} and the ANN-to-SNN conversion method\cite{kim2022privatesnn,rathi2020enabling} have been proposed.
The ANN-to-SNN conversion method directly obtains the weights of an SNN from a corresponding ANN, relying on the relationship between the firing rates of the SNN and activations of the ANN\cite{meng2022training}. 
The method enables the obtained SNN to obtain accuracy comparable to that of ANNs.
However, its latency is usually intolerably since only a large number of time steps can close the firing rates to activations of the ANN.
The significant latency also eliminates the potential of SNNs for high energy efficiency.
The significant latency also eliminates the potential of SNNs for high energy efficiency and fast response, making it hard to utilize in edge computing, etc.
On the other hand, surrogate gradient (SG) makes the non-differentiability SNNs can be trained directly with low latency\cite{xu2022ultra, xu2023ultra, zhang2019spike, zhou2023spikingformer} but cannot achieve high performance comparable to leading ANNs.
Insufficient performance also restricts its application in high-precision recognition fields such as intelligent healthcare, autonomous driving, etc.
Overall, it's still a challenge for traditional SNNs to achieve high accuracy, low latency, and energy efficiency for real-world applications.

In this paper, we overcome both the low performance and high latency issues of traditional SNNs by introducing spatio-temporal conversion blocks (STCBs) to propose a high energy efficiency spike-train level spiking neural network.
In traditional spiking convolutional neural networks (SCNNs), Integrate-and-Fire(IF) models or Leaky-Integrate-and-Fire(LIF) models are used to replace the ReLU function to make the information can be transmitted and processed in spiking form.
Non-differentiable spikes make SNNs challenging to train but have excellent potential for high energy efficiency.
Our basic idea is to utilize the advantages of spikes in energy consumption and keep the high performance of convolutional neural networks (CNNs).
The SNN hardware can utilize a binary spike-based sparse processing over a fixed number of time steps via accumulate (AC) operations, consuming much lower power than multiply-accumulate (MAC) operations that dominate computations in ANNs\cite{kundu2021towards}.
Convolutional operations consist of a large number of MAC operations, making them one of the primary sources of power consumption for convolutional neural networks\cite{kundu2021towards,datta2022can}.
So, we proposed spatio-temporal conversion blocks based on IF models to encode the real value of information into spike trains before convolutional layers, which can help reduce the energy consumption of convolution operation significantly.
To maintain the efficiency of information transmission between blocks, we use the current accumulator blocks (CABs) to decode the information after the convolutional layers.
To train our proposed STCSNNs with low latency effectively, we further study the error backpropagation of STCBs and deduce the equivalent gradient of STCBs.
We can train a high-performance, low-latency, and energy-efficiency STCSNN with these methods.

The following summarizes our main contributions:
\begin{itemize}
	\item We proposed high energy efficiency spike-train level spiking neural networks, which utilize the advantages of spikes in energy consumption and keep the high performance of convolutional neural networks (CNNs) meanwhile.
	\item We studied the error backpropagation of STCBs and proposed the corresponding equivalent gradient, which avoids the non-differentiability problem in SNN training and does not require the costly error backpropagation through the temporal domain.
	\item Our proposed STCSNN achieves competitive or state-of-the-art(SOTA) SNN performance with low latency and energy efficiency on the Fashion-MNIST\cite{xiao2017fashion}, CIFAR-10\cite{krizhevsky2009learning}, CIFAR-100\cite{krizhevsky2009learning}, TinyImageNet\cite{hansen2015tiny}, and DVS-CIFAR10\cite{li2017cifar10} datasets.
\end{itemize}

\section{Related work}

In recent years, SNNs have attracted widespread research interest and have developed rapidly.
However, there are still lots of challenging problems that remain to be unsolved.
Many works seek brain-inspired methods with biological plausibility to train SNNs\cite{caporale2008spike, kheradpisheh2018stdp,hebb2005organization}.
However, these methods cannot obtain competitive performance and are hard to apply to complicated datasets\cite{meng2022training}.
Besides the brain-inspired method, SNN training has been carried out in two strategies: ANN-to-SNN conversion and direct
training.
\paragraph{ANN-to-SNN conversion} The basic idea of ANN-to-SNN conversion is to convert a high-performance ANN to an SNN and adjust its parameters.
This method can avoid training problems of SNNs due to the non-differentiability.
But it usually needs high latency to make the firing rates closely approach the high-precision activation, which eliminates the potential of SNNs on energy efficiency.
To solve this problem, various technologies are proposed to reduce the latency, such as spike-norm \cite{sengupta2019going},  robust normalization \cite{rueckauer2016theory}, and channel-wise normalization \cite{kim2020spiking}.
Recently, Kim et al. proposed a novel attention-guided ANN compression method to convert compressed deep SNN models with high energy efficiency and reduced inference latency\cite{kundu2021towards}.
Li et al. proposed to apply adaptive threshold and layer-wise calibration to obtain high-performance SNNs\cite{li2021free}.
However, The SNNs obtained by the ANN-to-SNN conversion method have more significant latency than those SNNs trained directly.
In addition, the ANN-to-SNN convert method is not suitable for neuromorphic datasets.
In recent works about the ANN-to-SNN conversion method, researchers are trying to reduce the time steps of ANN-to-SNN conversion method and have achieved excellent performance with few time steps \cite{qu2024spiking,hu2023fast,guo2023transformer,you2024converting}.
\paragraph{Direct training} To solve the non-differentiability problem of SNNs, Lee et al. proposed to use a surrogate gradient to replace the non-differentiable activation term.\cite{lee2016training}.
Compared with the ANN-to-SNN conversion method, direct training achieves high accuracy with fewer time steps but suffers more training costs\cite{deng2020rethinking}.
Many studies use surrogate gradient (SG) to obtain high-performance SNNs on static and neuromorphic datasets\cite{xu2022ultra, xu2023ultra, zhang2019spike, li2021differentiable}.
Wu et al. first proposed the STBP method, which significantly improves the performance of SNNs obtained by direct training \cite{wu2018spatio}.
Zhang et al. proposed TSSL-BP to improve temporal learning precision by breaking down error backpropagation across two types of inter-neuron and intra-neuron dependencies and achieve low-latency and high-accuracy SNNs \cite{zhang2020temporal}.
Recently, Xu et al. proposed a novel training method based on backpropagation (BP) for SNNs with multi-threshold, which can reduce the latency of SNN to 1-2 time steps\cite{xu2022direct}.

\section{Methodology}
\subsection{Spatio-temporal conversion block (STCB)}
\label{section:STCB}
\begin{figure}[t]
\centerline{\includegraphics[width=\columnwidth]{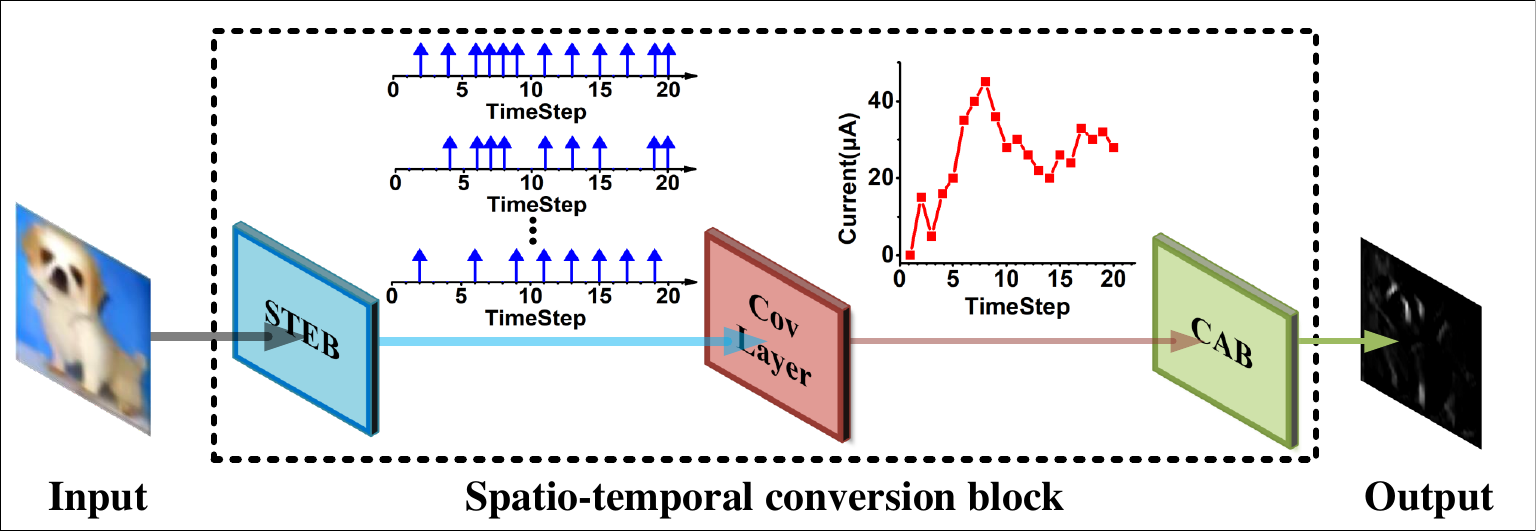}}
\caption{Spatio-temporal conversion block (STCB)}
\label{ST_Converasion}
\end{figure}

To reduce the energy consumption of convolutional layers, we proposed spatio-temporal conversion blocks (STCBs) to reduce the MAC operation of convolutional layers.
The STCB consists of a spike train encoding block (STEB), a convolutional layer, and a current accumulator block (CAB), which is shown in  Fig.\ref{ST_Converasion}.
The STCB utilizes IF models with decaying threshold voltage to encode information into spike trains, converting MAC operations of convolutional layers to AC operations.
Due to the feature of Integrate-and-Fire (IF) models, the STEB has a similar function to ReLU layers.
In STCB, we treat the convolutional layer as a synaptic model, which receives spikes and generates synaptic current responses.
The CAB is proposed to accumulate synaptic current with time steps, and the final accumulated synaptic current will be the output of the next STCB.
In STCB, information is transformed into binary information in the form of spike trains to simplify the convolutional operation and is recovered back to real information after convolutional operations.

\subsection{Spike train encoding block(STEB) and Current accumulator block (CAB)}
\label{section:STEB_CAB}

\begin{figure*}[!t]
{
\centering
\subfigure[]{
\includegraphics[width=0.7\columnwidth]{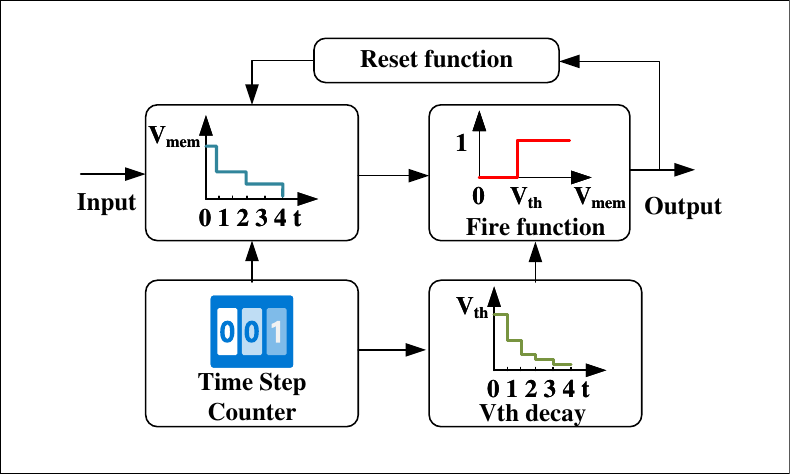}
\label{STEB}
}
\subfigure[]{
\includegraphics[width=0.7\columnwidth]{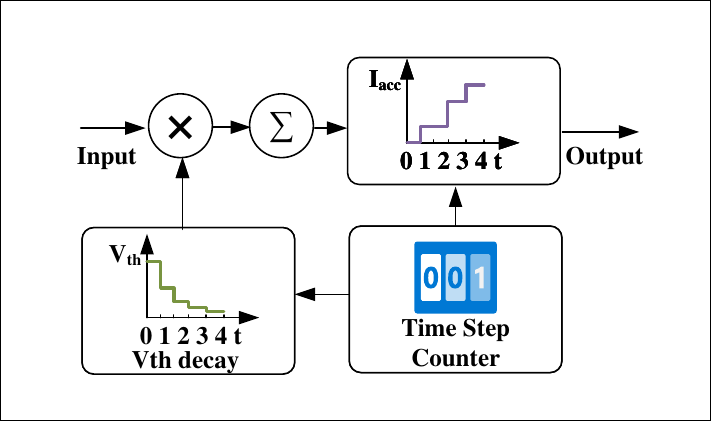}
\label{CAB}
}

\caption{The structure of STEB and CAB. (a) STEB (b) CAB}
}
\end{figure*}

\paragraph{Spike train encoding block(STEB)} As one of the main components of the STCBs, the spike train encoding block (STEB) utilizes the IF model\cite{burkitt2006review} to encode information into spike trains.
As Fig. \ref{STEB} shows, the input data will be used as the initial membrane voltage of the neuron model at the zeroth time step.
It is known that the IF model is one of the most widely applied models to describe neuronal dynamics in SNNs.
In this paper, we apply the IF models with decaying threshold voltages, whose threshold voltages decay with time steps, to encode input data into spike trains.
The IF model can be depicted as
\begin{equation}
    I(t)=C_m\frac{dV_m(t)}{dt},
\label{eq1}
\end{equation}
where $C_m$ is the membrane capacitance, $V_m(t)$ is membrane potential and $I(t)$ denotes the membrane current.
Since we only use input data $In[t]$ as the initial membrane, the neuronal membrane potential of neuron $i$ at time step $t$ is 

\begin{equation}
{u_i}[t] = \left\{ {\begin{array}{*{20}{r}}
{In[t],}& {t=0}\\
{u_i[t-1]+u_{reset}[t],}&{t>0}\\
\end{array}} ,\right.
\label{eq2}
\end{equation}
where $u_{reset}[t]$ denotes the reset function, which reduces the membrane potential by the $V_{th}[t]$ after the neuron $i$ fires at the $t$ time step. $u_{reset}[t]$ is equal to $-s_i[t]V_{th}[t]$. $s_i[t]$ is the output spike of neuron $i$ and $V_{th}[t]$ is the threshold voltage at time step $t$, which is given by

\begin{equation}
{s_i}[t] = \left\{ {\begin{array}{*{20}{r}}
{0,}& {u_i[t]<V_{th}[t]}\\
{1,}&{u_i[t]\geq V_{th}[t]}\\
\end{array}} ,\right.
\label{eq3}
\end{equation}

\begin{equation}
V_{th}[t] = \left\{ {\begin{array}{*{20}{r}}
{V_{init},} & {t=0}\\
{\frac{V_{th}[t-1]}{\tau_{vth}},} & {t>0}\\
\end{array}}, \right.
\label{eq4}
\end{equation}
where $V_{init}$ is the initial threshold voltage and $\tau_{vth}$ is the time constant of threshold voltages.

\paragraph{Current accumulator Block (CAB)} 
In STCBs, STEBs utilize IF models to encode information into spike trains.
As we mentioned in section \ref{section:STCB}, the convolutional layer is treated as synaptic model.
So, the spike train is converted into a series of synaptic currents after the convolutional operation.
We use the current accumulator block to accumulate the synaptic currents into a final synaptic current to improve the transmission efficiency between blocks.
Using CABs, we can transmit the information one time rather than transmit a spiking train many times between layers, which is more energy efficient in hardware design.
Because CABs aim to improve the efficiency of information transmission between network layers, we do not want to lose or increase extra information in CABs, which may decrease the network accuracy.
So, we utilize the reverse operation of STEBs as the operation of CABs to decode the information.
As Fig. \ref{CAB} shows, input data at time step $t$ is multiplied by the threshold voltage at time step $t$, and the result will be accumulated.
The output of CAB $I_{acc}[t]$ is calculated by

\begin{equation}
I_{acc}[t]=\left\{ {\begin{array}{*{20}{r}}
{0,} & {t<T_{max}}\\
{i_{acc}[t],} & {t=T_{max}}\\
\end{array}}, \right.
\label{eq5} 
\end{equation}
where $T_{max}$ is the maximum time step, which is also equal to the length of the spike trains. $i_{acc}[t]$ is the synaptic current at time step $t$, which is given by

\begin{equation}
i_{acc}[t]=\left\{ {\begin{array}{*{20}{r}}   
{I_{syn}[t]V_{th}[t],} & {t=0}\\
{i_{acc}[t-1]+I_{syn}[t]V_{th}[t],} & {t>0}\\
\end{array}}, \right.
\label{eq6} 
\end{equation}
where $I_{syn}[t]$ is the synaptic current at time step $t$, and $V_{th}[t]$ is the threshold voltage at time step $t$ which can be obtained by Eq. \ref{eq4}.

\subsection{Error Backpropagation of STCB}

The information is transmitted in the form of spike trains in STCB, while transmitted in the form of the accumulated synaptic current between STCBs, which makes error backpropagation of STCB complex.
To solve this issue, we proposed the equivalent gradient of STCBs, which can simplify the error backpropagation of STEB and reduce the computation.
As we mentioned in section \ref{section:STEB_CAB}, data will be first encoded into spike trains by STEB in the forward pass process.
Then, the spike trains will convert into a series of synaptic currents by the convolutional layer.
Finally, these synaptic currents will be accumulated by CAB, and the final accumulated synaptic current will be used as the output of STCB.
\begin{figure}[t]
\centerline{\includegraphics[width=0.8\columnwidth]{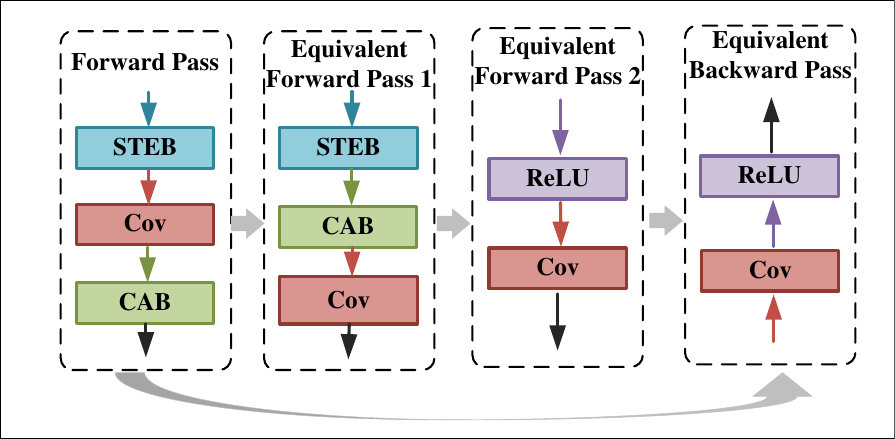}}
\caption{Forward and backward pass of STCB}
\label{BP}
\end{figure}
As Fig. \ref{BP} shows the equivalent forward pass 1 of STCB is obtained by reordering the Cov and CAB.
The forward calculation of Cov and CAB is 

\begin{equation}
O_{Cov+CAB}=I_{acc}[t]=\left\{ {\begin{array}{*{20}{r}}
{0,} & {t<T_{max}}\\
{i_{acc}[t]=i_{acc}[t-1]+I_{syn}[t]V_{th}[t],} & {t=T_{max}}\\
\end{array}}, \right.
\label{eqr1}
\end{equation}
where $O_{Cov+CAB}$ is the output of Cov and CAB.
Based on Eq. \ref{eq4}, Eq. \ref{eq5}, and Eq. \ref{eq6}, the forward calculation is simplified as

\begin{equation}
O_{Cov+CAB}==\left\{ {\begin{array}{*{20}{r}}
{0,} & {t<T_{max}}\\
{\sum_{k=0}^{T_{max}}Cov(S[T_{max}-k])\frac{V_{init}}{\tau_{vth}^{(T_{max}-k)}},} & {t=T_{max}}\\
\end{array}}, \right.
\label{eqr2}
\end{equation}
where $Cov$ is a convolution operation without bias, and $S[T_{max}-k]$ is the output of STEB at time step $[T_{max}-k]$, a matric composed of $s_i[T_{max}-k]$. The output of equivalent forward pass 1 of STCB is 

\begin{equation}
O_{CAB+Cov}=\left\{ {\begin{array}{*{20}{r}}
{0,} & {t<T_{max}}\\
{Cov(\sum_{k=0}^{T_{max}}S[T_{max}-k]\frac{V_{init}}{\tau_{vth}^{(T_{max}-k)}}),} & {t=T_{max}}\\
\end{array}} \right.
\label{eqr3}
\end{equation}

Based on the distributive law of convolution operation, we can see that the output of forward pass $O_{Cov+CAB}$ and equivalent forward pass 1 $O_{CAB+Cov}$ are the same.
In a neural network, the BP algorithm updates the network's weights by propagating the error backward. This process utilizes the chain rule to calculate gradients for adjusting each weight appropriately. Due to the chain rule holding regardless of the order of the functions, we simplify backward pass calculation by changing the order of forward pass calculations.
Based on Eq. \ref{eq3}, Eq. \ref{eq4}, Eq. \ref{eq5}, and Eq. \ref{eq6}, STEB and CAB are a spiking encoding and decoding process, respectively.
In STEBs, the input data larger than zero will be quantified to a spiking train based on the threshold voltage $V_{th}[t]$, and the input data less than zero will be encoded to zero. 
In CABs, the spiking trains will be recovered to synaptic current based on the threshold voltage $V_{th}[t]$.
During the quantization process in STEBs, due to the limited length of the spiking trains, there are some quantization errors.
If we ignore the quantization error of the encoding process of STEB, STEB + CAB is equal to ReLU.
So, we can obtain an approximate equivalent forward pass 2.
We can use the backward pass of ReLU + Cov to replace the one of STCB, which can simplify the error backpropagation of STCB and significantly reduce the computation.

In our proposed SNN, the original forward pass STEB+Cov+CAB is used as a forward calculation to preserve SNNs' advantages during the training and testing process, and the equivalent backward pass ReLU+ Cov is used as a backward calculation to simplify calculating gradients during the training process.
In other words, the model STEB+Cov+CAB is used in the forward calculation process, and the error is calculated based on the loss function. ReLU+Cov is used in the backward calculation process, and the error gradient with respect to the weight in each layer is calculated based on the gradient function of ReLU and Cov.

\subsection{Network Architecture}
\label{section:structure}
\begin{figure}[!t]
{
\centering
\subfigure[]{
\includegraphics[width=0.42\columnwidth]{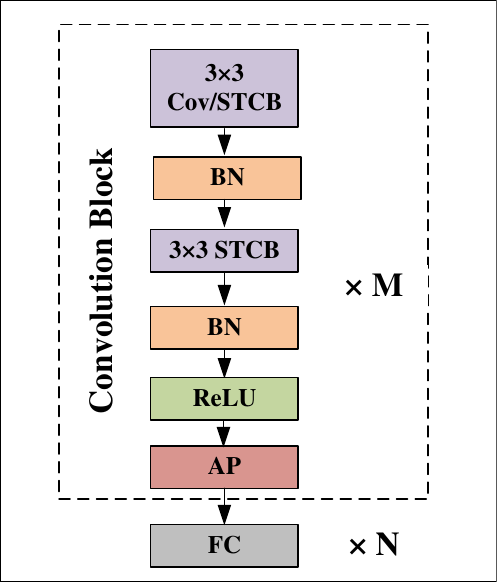}
\label{VGG}
}
\subfigure[]{
\includegraphics[width=0.5\columnwidth]{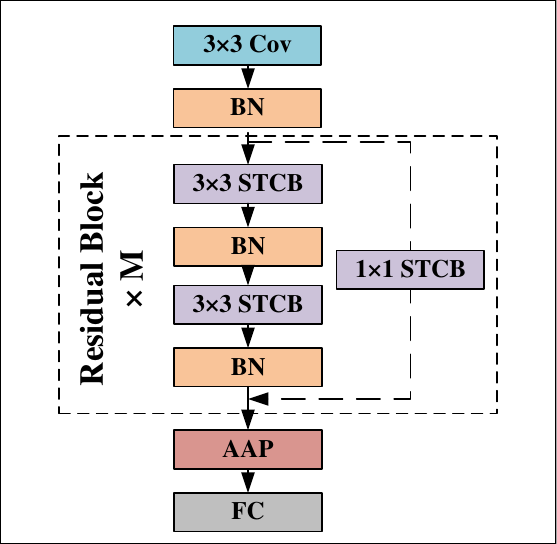}
\label{ResNet}
}

\caption{Diagram of VGG and ResNet based on STCB. (a) VGG, (b) ResNet. Cov refers to the convolutional layer, $N\times N$ Cov/STCB means the Convolutional layer or STCB with $N \times N$  filters, BN means Batch Norm, AP is the average pooling layer, AAP is the adaptive average pool, and FC refers to the fully connected layers.}
}
\end{figure}

Using the STCB mentioned in the previous section, we propose an STCSNN that accurately classifies static and neuromorphic datasets with a low computational cost and energy consumption.
In this section, we chose VGG and ResNet architecture to show how to build an STCSNN.
Fig. \ref{VGG} and \ref{ResNet} show the diagram of VGG and ResNet based on STCBs.
For the first convolutional block of VGG architecture, we keep a convolutional layer with 3$\times$3 filters as the first layer, which helps to obtain enough features from the original data.
In other convolution blocks, we just need to use the STCB to replace the convolutional layers. 
For ResNet architecture, we also keep a convolutional layer with 3$\times$3 filters as the first layer.
Due to the feature of IF models, the STEB also has the function of ReLU layers.
So, we use the STCB to replace the convolutional layer + ReLU function in residual blocks.

\section{Experiments and results}
In this section, we first evaluate the proposed STCSNN on four static datasets, Fashion-MNIST\cite{xiao2017fashion}, CIFAR-10\cite{krizhevsky2009learning}, CIFAR-100\cite{krizhevsky2009learning}, TinyImageNet\cite{hansen2015tiny}, and one and neuromorphic dataset, DVS-CIFAR10\cite{li2017cifar10} with different structures of SNNs.
We compared our method with several previously reported state-of-the-art results with the same or similar networks, including different SNNs trained by BP-based methods and converted SNNs.
We also study the influence of initial threshold voltages and the length of the spike trains to show the tradeoff between accuracy and latency.
Finally, we count the number of spikes for different datasets and analyze the energy consumption of STCSNNs.

\subsection{Experiment settings}
All reported experiments below are conducted on an NVIDIA Tesla V100 GPU.
The implementation of our proposed method is on the Pytorch framework \cite{paszke2019pytorch}.
The experimented SNNs are based on the network structure described in Sec. \ref{section:structure}.
Only 8 time steps are used to demonstrate the proposed ultra low-latency spiking neural network.
No refractory period is used.
SGD\cite{rumelhart1986learning} is applied as the optimizer.
If not otherwise specified, the accuracy in this paper refers to the best results obtained by repeating the experiments five times.
The detailed experiment setting is presented in Table \ref{T1}.
For static datasets, the input is treated as the initial membrane potential in STEB.
For neuromorphic datasets, the event-to-frame integrating method \cite{fang2021incorporating} is used, and the integrated frame is treated as the initial membrane potential in STEB.
In addition, we do not apply complex skills, such as error normalization\cite{diehl2015fast}, weight regularization\cite{lee2016training},
warm-up mechanism\cite{zhang2020temporal}, etc

\begin{table*}[h]
\caption{Experiment settings}
\label{T1}
\centering
\resizebox{0.8\textwidth}{!}
{
\begin{tabular}{lll} 
\hline
Parameters         & Description                               & Value                 \\ 
\hline
\textit{$V_{init}$}       &    \begin{tabular}[c]{@{}l@{}}Initial threshold voltage\\(FashionMnist, Cifar10/Cifar100/\\TinyImageNet/DVS-Cifar10)\end{tabular}                                                      &  \begin{tabular}[c]{@{}l@{}}0.8mV, 1mV, \\2mV, 4mV \end{tabular}\\
\textit{$L$} & \begin{tabular}[c]{@{}l@{}}length of spike trains which is equal\\ to the maximum time step $T_{max}$ \end{tabular} & 8\\
\textit{$\tau_{vth}$} & time constant of threshold voltages & 2\\
\textit{$N_{Batch}$}    & Batch Size         & 32               \\
\textit{$\eta$}         & \begin{tabular}[c]{@{}l@{}}Learning rate with cosine decay to 0\\(FashionMnist, Cifar10/Cifar100,\\ TinyImageNet, DVS-Cifar10)\end{tabular} & \begin{tabular}[c]{@{}l@{}}0.001, 0.005,\\ 0.01, 0.005 \end{tabular} \\
\textit{$epoch$}         & \begin{tabular}[c]{@{}l@{}}Learning rate\\(FashionMnist, Cifar10/Cifar100,\\ TinyImageNet/DVS-Cifar10)\end{tabular} & 100, 300, 200 50  \\
\hline
\end{tabular}
}
\end{table*}

\subsection{ Datasets and training details}
\paragraph{FashionMnist} The FashionMNIST dataset \cite{xiao2017fashion} of clothing items contains 60,000 labeled training images and 10,000 labeled testing images, each of which is also a $28 \times 28$ grayscale image, as with MNIST. Compared with MNIST, FashionMNIST is a more challenging dataset that can serve as a direct drop-in replacement for the original MNIST dataset. 
We use ResNet18 for FashionMnist. 
Moreover, the random horizontal crop is applied to the training images.
The learning rate decays to 50\% every 20 epochs.

\paragraph{Cifar10 and Cifar100} The CIFAR dataset \cite{krizhevsky2009learning} consists of 50,000 training images and 10,000 testing images with the size of 32 × 32.
We apply VGG16 and ResNet18 for both CIFAR10 and CIFAR100. 
Moreover, the random horizontal flip and crop are applied to the training images as data augmentation.
We apply a learning rate with cosine decay to 0.

\paragraph{TinyImageNet} TinyImageNet\cite{hansen2015tiny} contains 100,000 training images of 200 classes (500 for each class) and 10,000 testing images, which are downsized to 64×64 colored images.
We still apply VGG16 and ResNet18 for both TinyImageNet. 
Moreover, the random horizontal flip and crop are applied to the training images as data augmentation.
We apply a learning rate with cosine decay to 0.

\paragraph{DVS-Cifar10} Cifar10-DVS\cite{li2017cifar10} is a neuromorphic version converted from the Cifar10 dataset in which 10,000 frame-based images with the size $128 \times 128$ are converted into 10,000 event streams with DVS.
Following \cite{SpikingJelly}, we apply the event-to-frame integrating method for pre-processing DVS-Cifar10.
We randomly split the dataset into 9,000 training images and 1,000 test images.
No augmentation is applied.
We apply a learning rate with cosine decay to 0.

\subsection{Comparison with existing works}
The comparison results on FashionMnist, CIFAR-10, CIFAR-100, TinyImageNet, and DVS-CIFAR10 are shown in Tab. \ref{T2}.

For the FashionMnist dataset, we use ResNet18\cite{he2016identity} as the network structure.
Tab. \ref{T2} shows that the proposed STCSNN method outperforms all other methods on FashionMnist with 8 time steps, based on 5 runs of experiments. 
Although some direct training methods such as \cite{suetake2023s3nn} use smaller time steps than ours, our method can achieve better performances.
To show the performance change after we use our proposed STCB to replace the ReLu + Cov, we implement the ANN with the same structures and train it in the same experimental conditions.
For Fashion MNIST, though our proposed method has a lower mean accuracy, our proposed method got the same best accuracy.
For the FashionMnist dataset, which is a simple dataset, our proposed method still obtained 0.51\%-5.19\% accuracy improvement compared with other SNNs.

For the Cifar-10 and Cifar-100 datasets, we use VGG16\cite{kundu2021towards} and ResNet18\cite{he2016identity} as the network architectures.
On Cifar-10 and Cifar-100, our proposed STCSNN achieves the highest accuracy based on 5 runs of experiments compared with other directing trained SNNs.
Compared with the ANN-SNN method, our proposed method can obtain competitive or even better results.
For Cifar-10, Our proposed STCSNN based on VGG16 and ResNet18 architecture obtains $5.43\%$ and $95.96\%$ test accuracy with only 8 time steps, respectively.
Though \cite{suetake2023s3nn,kim2022neural,deng2022temporal} use fewer time steps, our work achieved $1.52\%-5.29\%$ accuracy improvement for Cifar100 with simpler network structure.

For the TinyImageNet dataset, we still used VGG16\cite{kundu2021towards} and ResNet18\cite{he2016identity} as the network architectures.
Compared to the ANN-to-SNN methods, which apply the same VGG16 architecture, our method obtains a higher accuracy with fewer time steps.
If we apply a more powerful architecture ResNet18, STCSNN can obtain $60.54\%$ accuracy, which improves the accuracy by $2.74\%$-$13.75\%$ compared with other methods.
For Cifar-10, Cifar-100, and TinyImageNet, our proposed method has no significant accuracy drop compared with ANN.

We also test our STCSNN on the neuromorphic DVS-CIFAR10.
We adopt the VGG-11\cite{rathi2020enabling} and VGG16\cite{kundu2021towards} architectures and conduct 5 runs of experiments for each architecture in 50 epochs.
As Tab. \ref{T2} shows, the proposed method outperforms other SOTA methods with low latency using VGG11 and VGG16.
Due to the ability to process spatiotemporal information of SNNs, our proposed method achieved better results on the neuromorphic DVS-CIFAR10 than ANN.

\begin{table*}
\caption{Compare with exciting work on different datasets.}
\label{T2}
\centering
\resizebox{1\textwidth}{!}{
\begin{tabular}{cccccccc} 
\hline
Dataset & Model &Methods &Architecture &Time Steps &mean(\%)&stddev(\%)&Best(\%)\\
\hline
\multicolumn{1}{c}{\multirow{4}{*}{Fashion MNIST}} &Zhang and Li.\cite{zhang2019spike} &SNN &1liner,1Recurrent &400 &90.0 & 0.14 &90.13\\ 
&Cheng, et al.\cite{cheng2020lisnn} &SNN &2Convs.1Linears  &20 &92.07 &- &-\\
&Xu, et al \cite{xu2022direct} &SNN &2Convs.1Linears &2  &- &- &93.80\\
&Fang, et al \cite{fang2021incorporating} &SNN &2Convs.1Linears &8  &- &- &94.38\\
&Kazuma Suetake et al.\cite{suetake2023s3nn} &SNN &ResNet18 &1 &- &- &94.81\\
&ANN  &ANN &ResNet18 &1 &95.18 &0.21  &95.32\\ 
&Our model &SNN &ResNet18 &8 &95.05 &0.26  &95.32\\ 
\hline
\multicolumn{1}{c}{\multirow{5}{*}{Cifar10}} &Kim, et al.\cite{kim2022privatesnn} &ANN-to-SNN &VGG16 &150 &- &- &89.2\\ 
&Zhang and Li\cite{zhang2020temporal} &SNN &5Convs. 2Linears &5 &- &-  &91.41\\ 
&Rathi, et al\cite{rathi2020enabling} &ANN-to-SNN &ResNet20 &250 &- &-  &92.22\\
&Kundu, et al\cite{kundu2021towards} &ANN-to-SNN &VGG16 &10 &- &- &92.53\\
&Deng, et al\cite{deng2022temporal} &SNN &ResNet19 &6 &94.50 &- &94.57\\
&Hao, et al.\cite{hao2023reducing} &ANN-SNN &VGG16 &8 &- &- &95.52\\
&Hao, et al.\cite{hao2023reducing} &ANN-SNN &ResNet18 &8 &- &- &95.60\\
&ANN &ANN &VGG16 &1 &95.56 &0.12  &95.83\\ 
&ANN &ANN &ResNet18 &1 &95.64 &0.18  &96.04\\ 
&Our model &SNN &VGG16 &8 &95.34 &0.19 &95.43\\ 
&Our model &SNN &ResNet18 &8 &95.79 &0.14 &95.96\\ 
\hline
\multicolumn{1}{c}{\multirow{6}{*}{Cifar100}} &Shen et al.\cite{shen2020balanced} &BNN &ResNet20 &1 &- &- &69.38\\ 
&Kim et al.\cite{kim2022privatesnn} &ANN-to-SNN &VGG16 &200 &- &-  &62.30\\ 
&Rathi et al.\cite{rathi2020enabling} &ANN-to-SNN &VGG11 &125 &- &-  &67.87\\
&Kazuma Suetake et al.\cite{suetake2023s3nn} &SNN &ResNet18 &1 &- &-  &71.23\\
&Kim, Li et al.\cite{kim2022neural} &SNN &NAS &5 &73.04 &-  &73.40\\
&Deng, et al\cite{deng2022temporal} &SNN &ResNet19 &6 &74.72 &- &75.00\\
&Hao, et al.\cite{hao2023reducing} &ANN-SNN &VGG16 &8 &- &- &76.25\\
&Hao, et al.\cite{hao2023reducing} &ANN-SNN &ResNet20 &8 &- &- &62.94\\
&ANN &ANN &VGG16 &1 &76.01 &0.12  &76.28\\ 
&ANN &ANN &ResNet18 &1 &76.28 &0.34  &77.01\\ 
&Our model &SNN &VGG16 &8 &75.68 &0.25 &76.05\\ 
&Our model &SNN &ResNet18 &8 &75.95 &0.18 &76.52\\ 
\hline
\multicolumn{1}{c}{\multirow{6}{*}{TinyImageNet}} &Na et al.\cite{na2022autosnn} &SNN &NAS &8 &- &- &46.79\\ 
&Kim et al.\cite{kim2022privatesnn} &ANN-to-SNN &VGG16 &200 &- &- &51.90\\ 
&Kundu et al.\cite{kundu2021towards} &ANN-to-SNN &VGG16 &10 &- &- &54.10\\
&Kim, Li et al.\cite{kim2022neural} &SNN &NAS &5  &- &- &54.60\\
&Kim and Panda.\cite{kim2021revisiting} &SNN &VGG11 &30  &- &- &57.80\\
&ANN &ANN &VGG16 &1 &58.32 &0.16  &58.63\\ 
&ANN &ANN &ResNet18 &1 &60.42 &0.26  &61.26\\ 
&Our model &SNN &VGG16 &8 &58.14 &0.21  &58.34\\
&Our model &SNN &ResNet18 &8 &59.99 &0.48  &60.54\\
\hline
\multicolumn{1}{c}{\multirow{6}{*}{DVS-Cifar10}} &Wu, et al.\cite{wu2021training} &SNN &VGG7 &50 &- &- &62.50\\ 
&Wu, et al.\cite{wu2021tandem} &SNN &7Convs.2Linears &20 &- &- &65.59\\ 
&Zheng, et al.\cite{zheng2021going} &SNN &ResNet19 &10 &- &- &67.80\\
&Xu, et al.\cite{xu2022ultra} &SNN &6Convs.1Linear & 5 &- &- &73.80\\
&Fang, et al.\cite{fang2021incorporating} &SNN &7Convs.2Linears &20 &- &- &74.80\\
&Meng, et al.\cite{meng2022training} &SNN &VGG11 &20 &75.03 &- &75.42\\
&Deng, et al\cite{deng2022temporal} &SNN &ResNet19 &10 &77.33 &- &77.54\\
&ANN &ANN &VGG11 &1 &74.72 &0.21  &75.10\\ 
&ANN &ANN &VGG16 &1 &76.20 &0.32  &75.60\\  
&Our model &SNN &VGG11 &8 &76.00 &1.11  &77.50\\
&Our model &SNN &VGG16 &8 &77.93 &0.21  &78.20\\
\hline
\end{tabular}
}
\end{table*}

\subsection{Performance analysis}

\begin{figure*}[!t]
{
\centering
\subfigure[]{
\includegraphics[width=0.5\columnwidth]{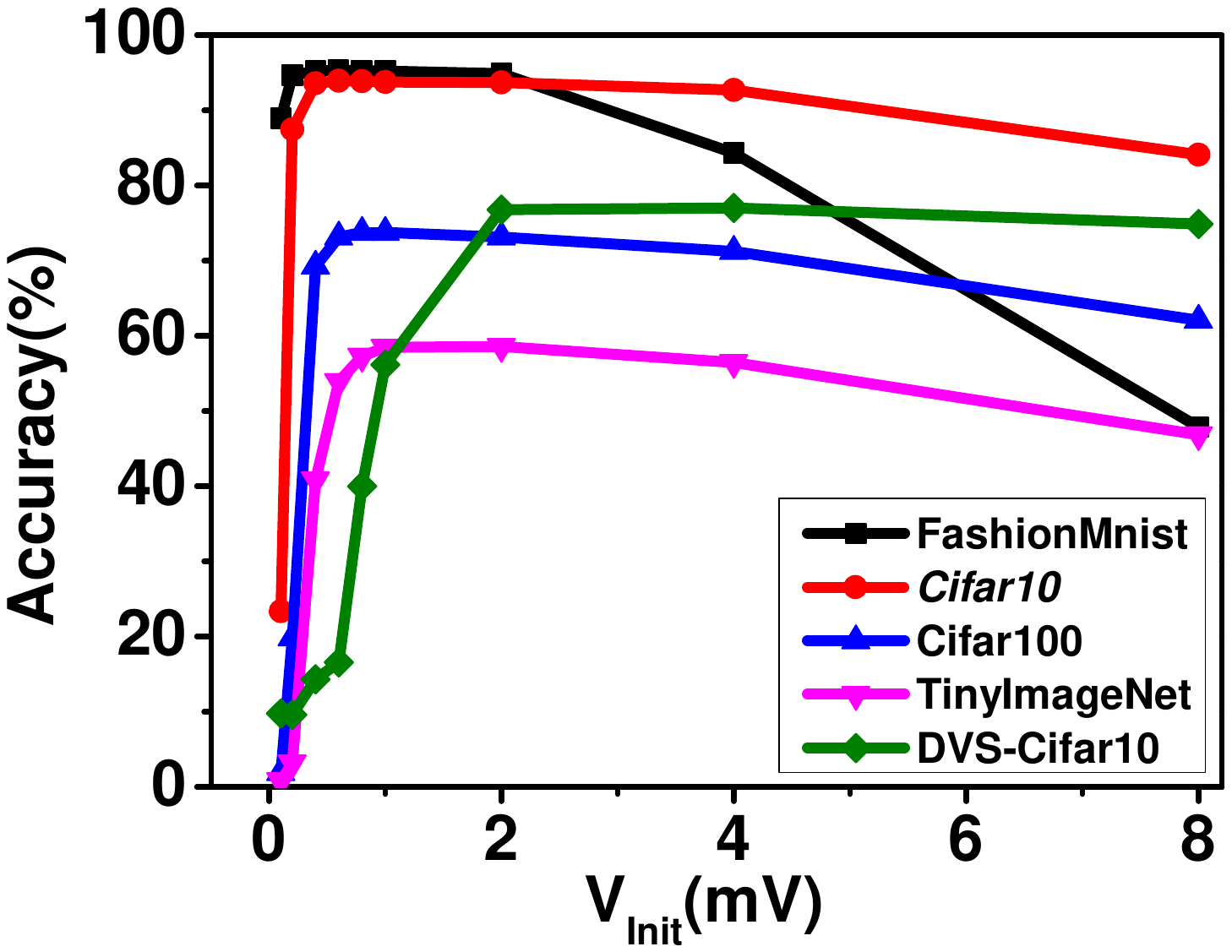}
\label{Vth_initial}
}
\subfigure[]{
\includegraphics[width=0.5\columnwidth]{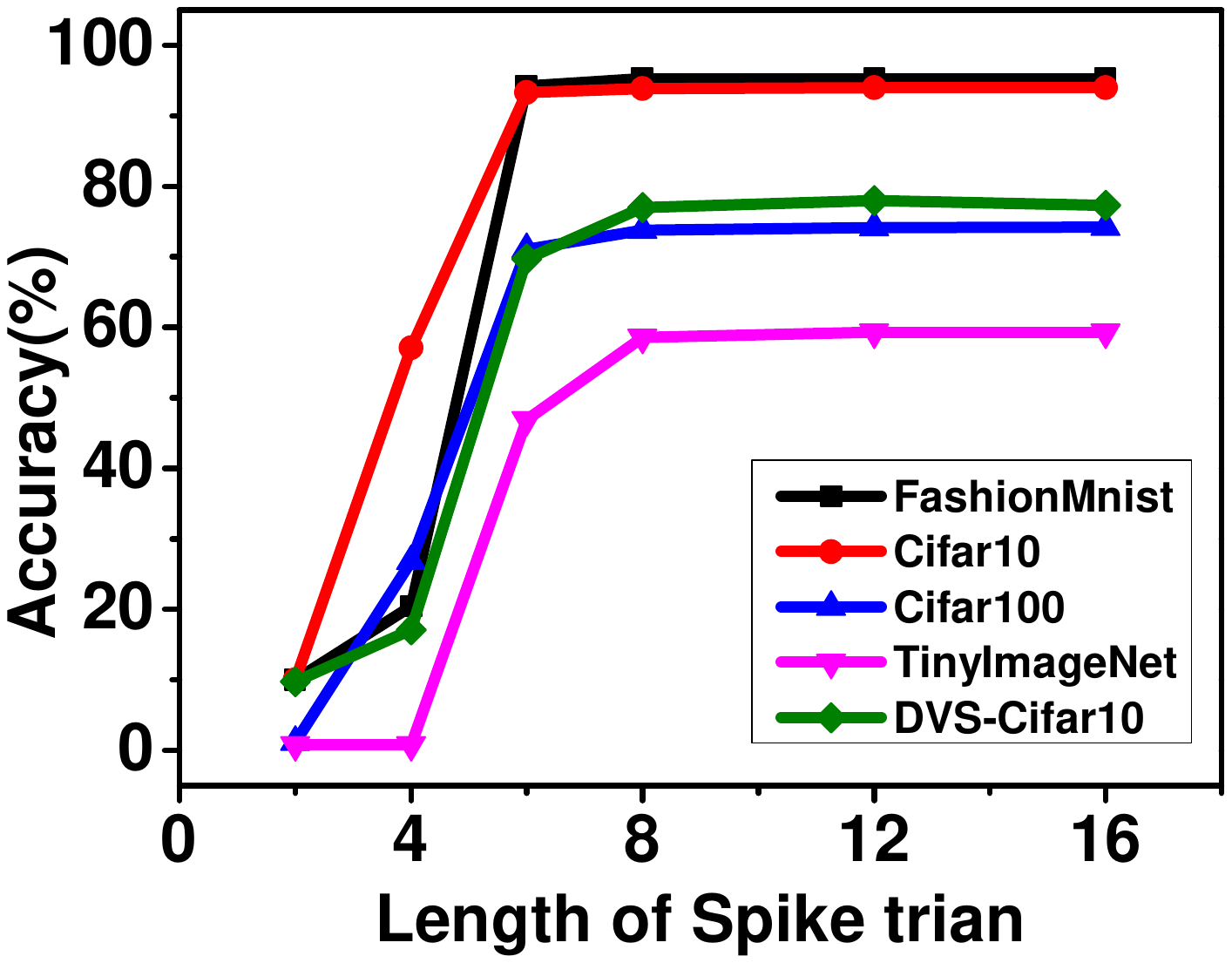}
\label{lengthofspike}
}

\caption{Performance analysis. (a) Initial threshold voltages (b) length of spike trains}
}
\end{figure*}

Since the initial threshold voltage and the length of the spike trains directly affect the loss of information during the encoding and decoding process in STCB, we study the influence of initial threshold voltages and length of spike trains on the accuracy of SNNs using FashionMnist, Cifar10, Cifar100, TinyImageNet, and DVS-Cifar10.
\subsubsection{Influence of initial threshold voltages}
\label{Influence of initial threshold voltages}
We chose different voltages (0.1, 0.2, 0.4, 0.6, 0.8, 1, 2, 4, 8mV) as the initial threshold voltage of STCSNNs.
For static datasets, FashionMnist, Cifar10, Cifar100, and TinyImageNet, we use ResNet18 as the network structure.
For the neuromorphic dataset DVS-Cifar10, VGG16 is applied.
As Fig. \ref{Vth_initial} shows, there is a significant accuracy drop when the initial threshold voltage is less than 0.8 mV.
STCSNN can obtain a reasonable accuracy for the static datasets when the initial threshold voltage is larger than 1.
We further study the distribution of the input of STCSNNs and find that the small initial threshold voltage may cause serious information loss during the encoding process in STCB.
For FashionMnist, Cifar10, Cifar100, and TinyImageNet datasets, the input of each STCB has a similar distribution and is in the range of [-1, 1].
When the initial threshold voltage is too small, such as 0.1, the part of the input larger than 0.1 will be encoded into one spike, which may cause profound information loss.
For DVS-Cifar10, the input of each STCB has a similar distribution and is in the range of [-2, 2].
So, the initial threshold voltage is less than 2, which will make a significant accuracy drop.

\subsubsection{Influence of the length of spike trains}
\begin{figure*}[!t]
{
\centering
\subfigure[]{
\includegraphics[width=0.7\columnwidth]{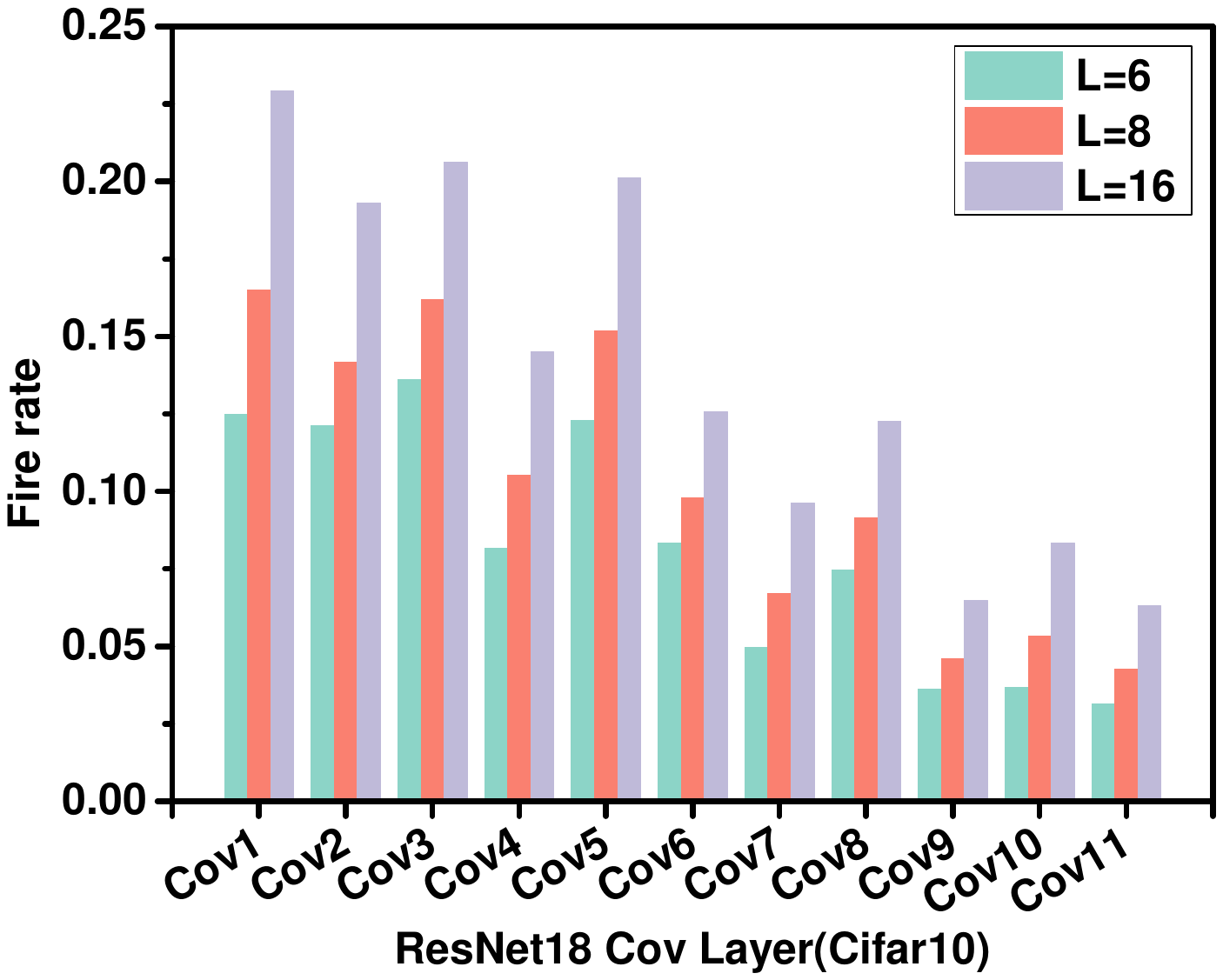}
\label{FireRate_Cifar10}
}
\subfigure[]{
\includegraphics[width=0.7\columnwidth]{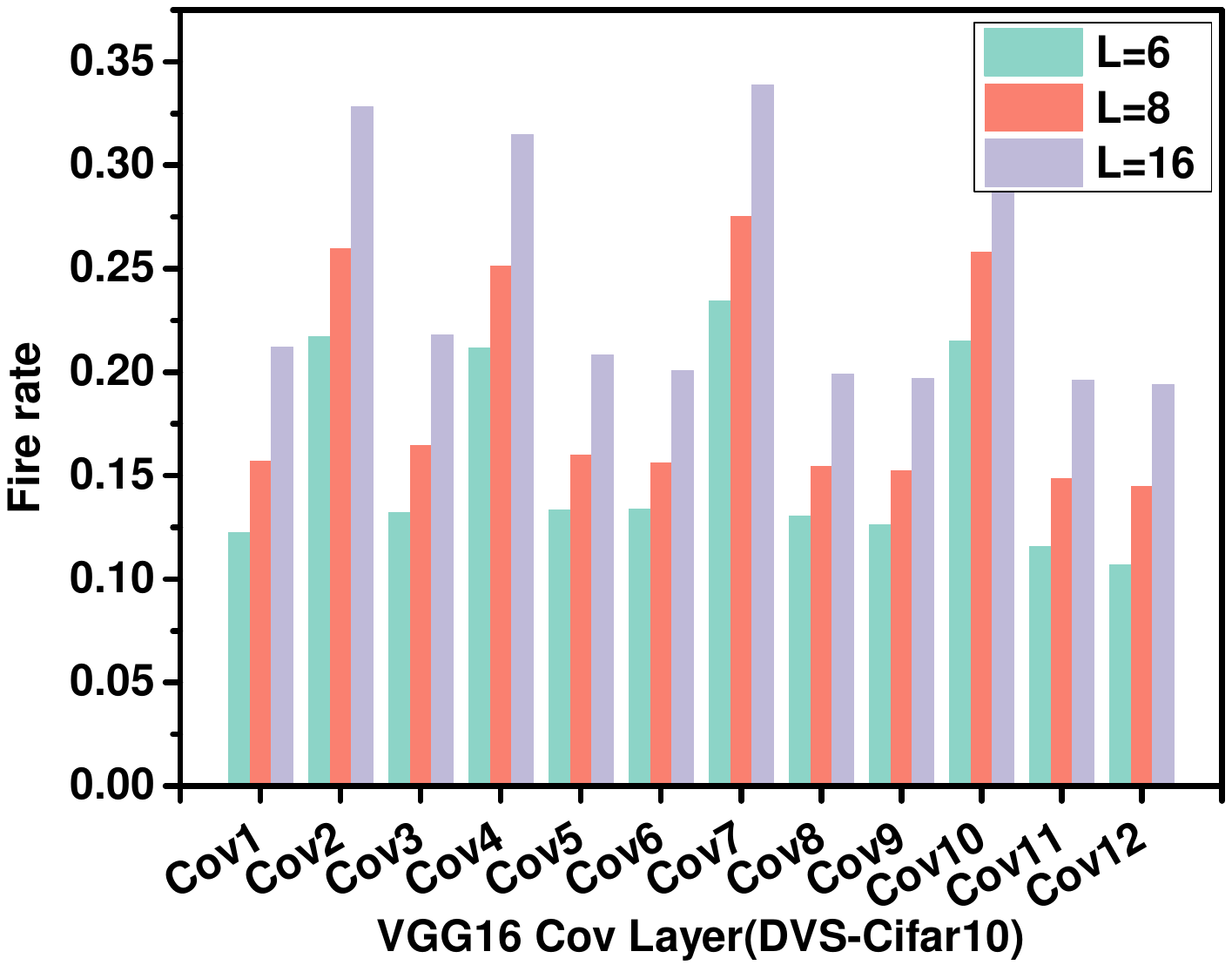}
\label{FireRate_DVSCifar10}
}
\caption{Average fire rate. (a) ResNet18 (Cifar10) (b)VGG16 (DVS-Cifar10)}
}
\end{figure*}

\begin{figure*}[!t]
{
\centering
\subfigure[]{
\includegraphics[width=0.7\columnwidth]{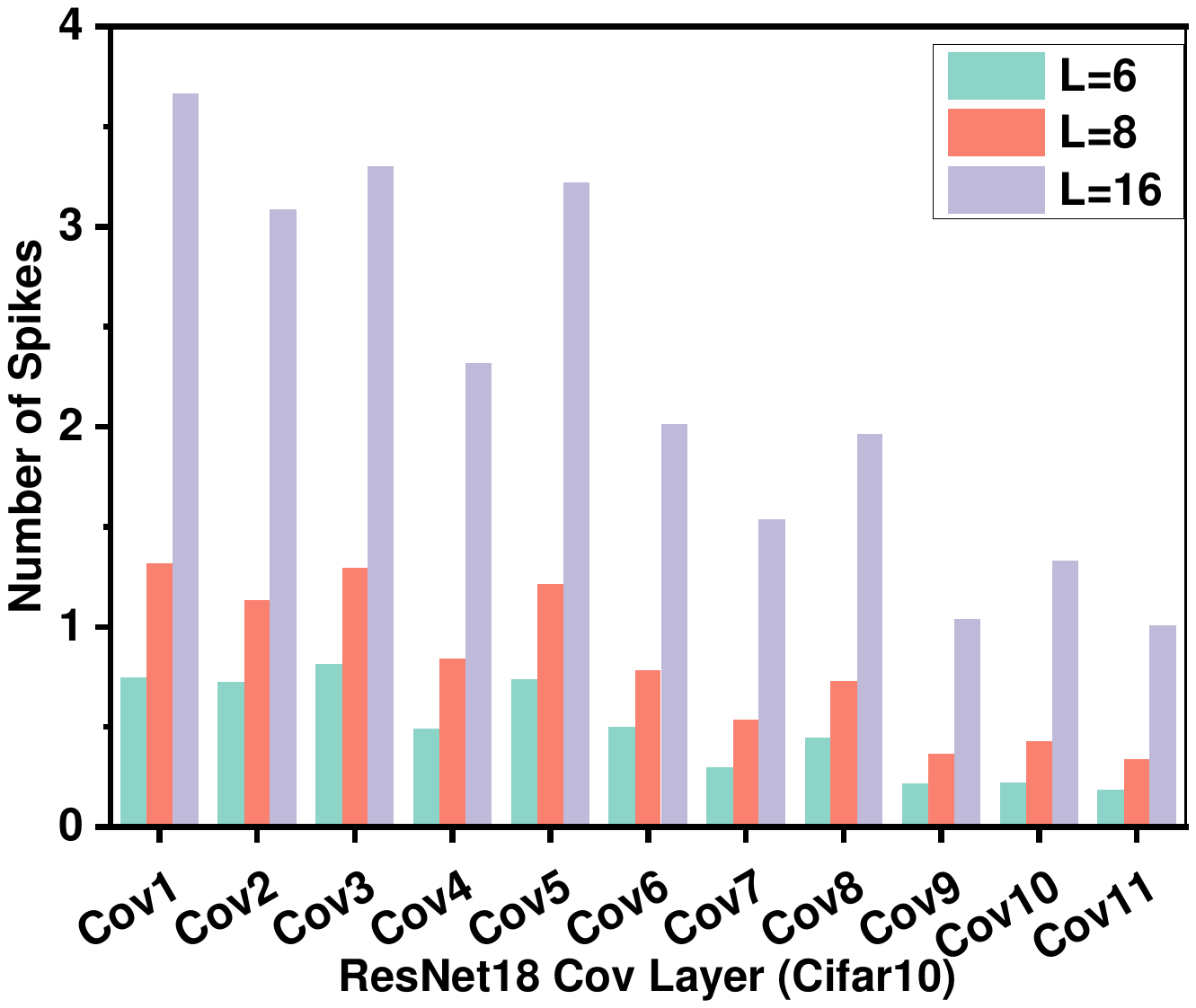}
\label{SpikeNum_Cifar10}
}
\subfigure[]{
\includegraphics[width=0.7\columnwidth]{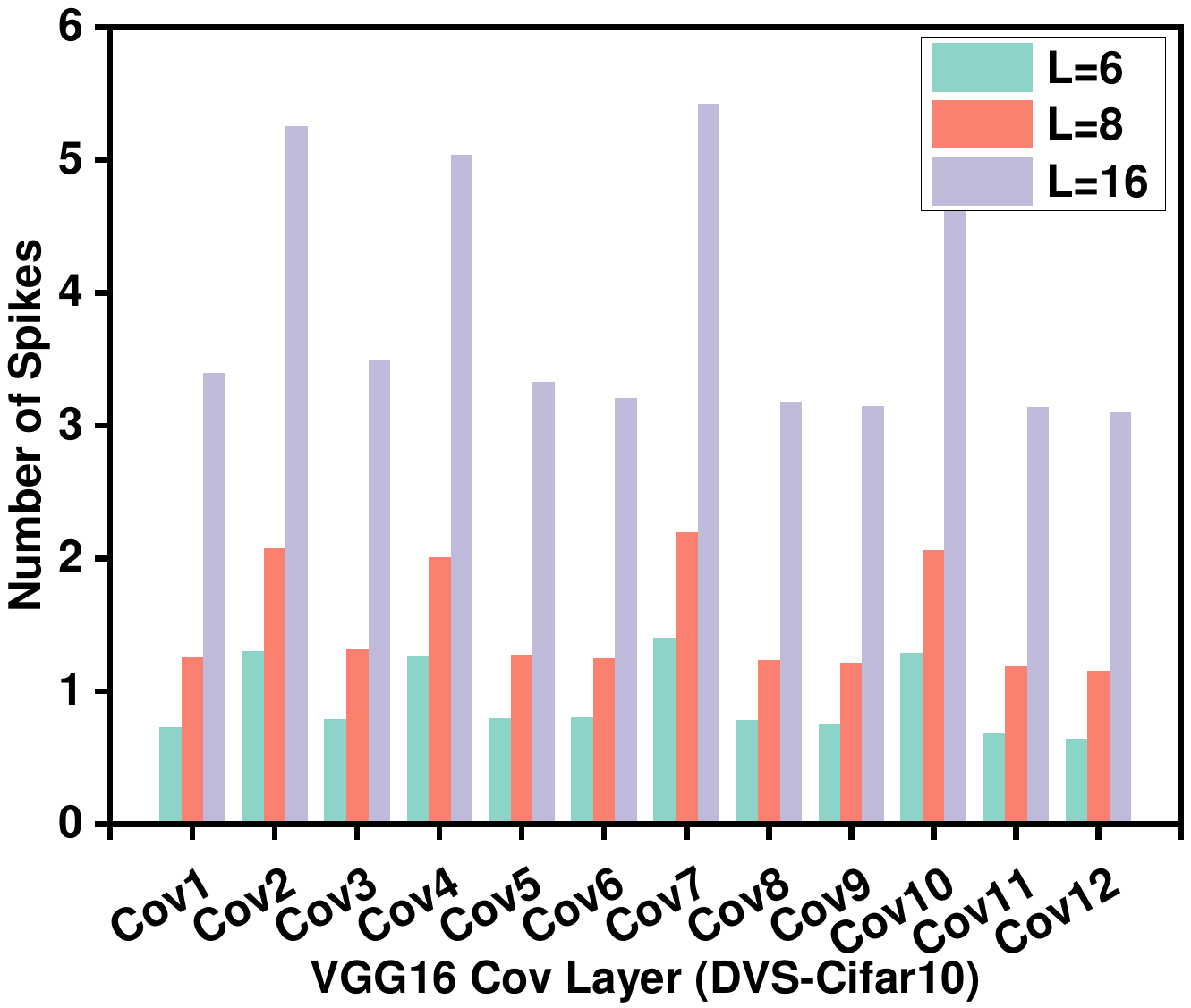}
\label{SpikeNum_DVSCifar10}
}
\caption{Average number of spikes. (a) ResNet18 (Cifar10) (b)VGG16 (DVS-Cifar10)}
}
\end{figure*}
We chose different lengths of spike trains (2, 4, 6, 8, 12, 16) to study the influence on the accuracy of STCSNNs.
The network architecture and datasets are the same as section \ref{Influence of initial threshold voltages}.
The initial threshold voltages are set as 0.8 and 2 for the static datasets (FashionMnist, Cifar10, Cifar100, and TinyImageNet datasets) and the neuromorphic dataset (TinyImageNet), respectively.
Fig. \ref{lengthofspike} is the accuracy of STCSNN with different lengths of spike trains on different datasets.
When the length of spike trains is less than 4, the STCSNN cannot work properly. 
For a fixed initial threshold voltage, the length of spike trains determines the encoding and decoding precision of STCB.
So, too short spike trains may cause the smaller input information to be lost, which causes STCSNN accuracy to drop significantly, especially for the deeper network.
Fig. \ref{lengthofspike} also shows that the accuracy of STCSNN is not significantly increasing with the increase of the length of the spike train when the length is larger than 8.
Longer spike trains mean more energy consumption.
Therefore, a suitable initial threshold voltage and length of the spike train can make a better tradeoff between accuracy and energy.

\subsection{Energy consumption analysis}
In this section, we first study the fire rate and the number of spikes of each convolutional layer of ResNets18 and VGG16. 
Then, we use FLOPs count to capture the computation energy to show the high energy efficiency of STCSNNs.

\subsubsection{Fire rate and Number of spikes}
We use Cifar10 and DVS-Cifar10 datasets to show the average fire rate and the number of spikes of each convolutional layer of ResNets18 and VGG16, respectively.
As Fig. \ref{FireRate_Cifar10} and \ref{FireRate_DVSCifar10}shows, the average fire rate increases with the increasing spike train's length. 
Fig. \ref{FireRate_Cifar10} depicts the benefits of the short length of spike trains, achieving significantly lower spike activity for almost all convolutional layers.
Fig. \ref{SpikeNum_Cifar10} and \ref{SpikeNum_DVSCifar10} show the average number of spikes of ResNet18 and VGG16.
Because the number of spikes is determined by the fire rate and the length of spike trains, the number of each convolutional layer increases rapidly with the increase of the length of spike trains.
The good news is that we do not need a very long spike train to obtain a competitive result.

\begin{figure*}[!t]
{
\centering
\subfigure[]{
\includegraphics[width=0.70\columnwidth]{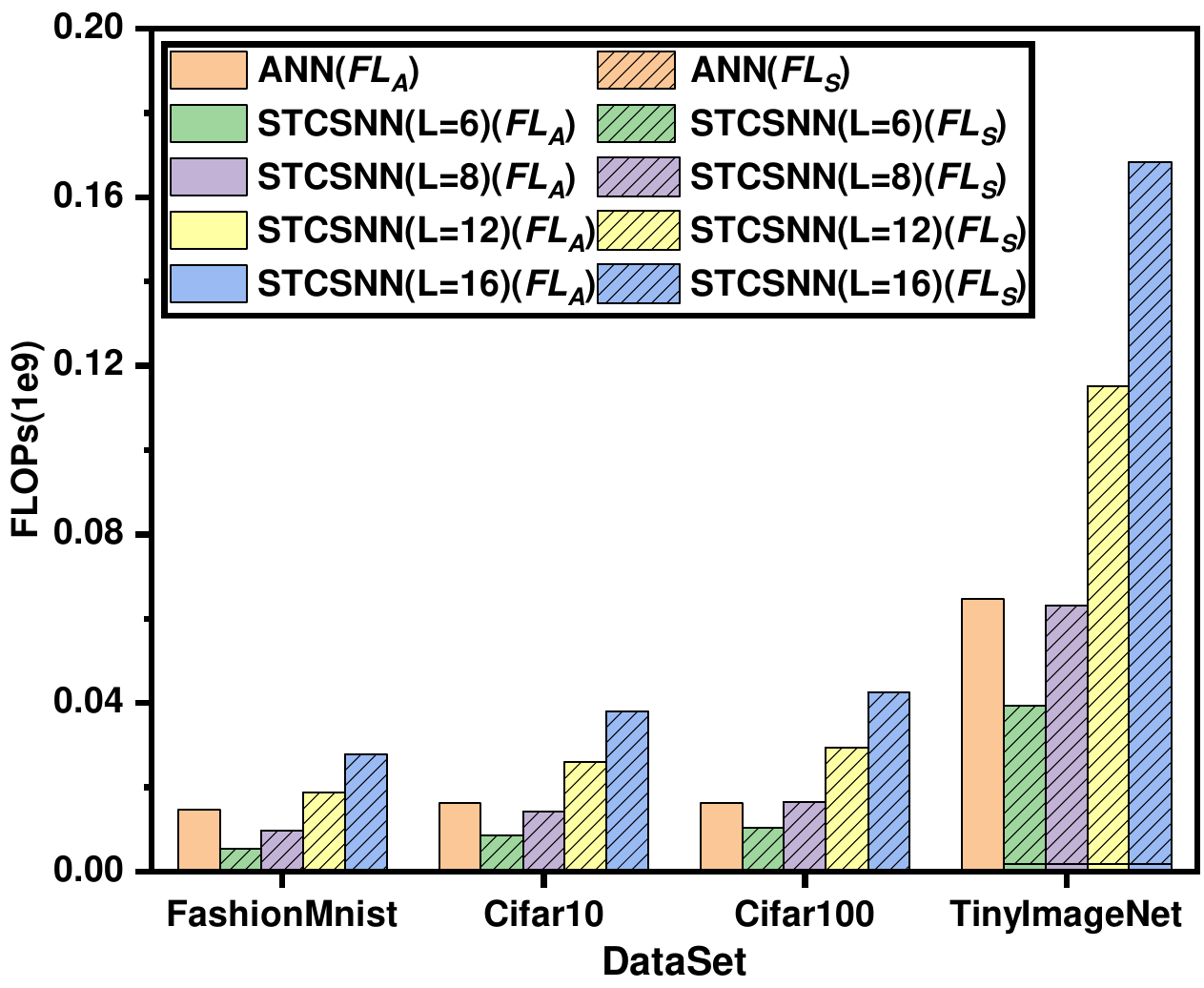}
\label{FLOPsStatic}
}
\subfigure[]{
\includegraphics[width=0.70\columnwidth]{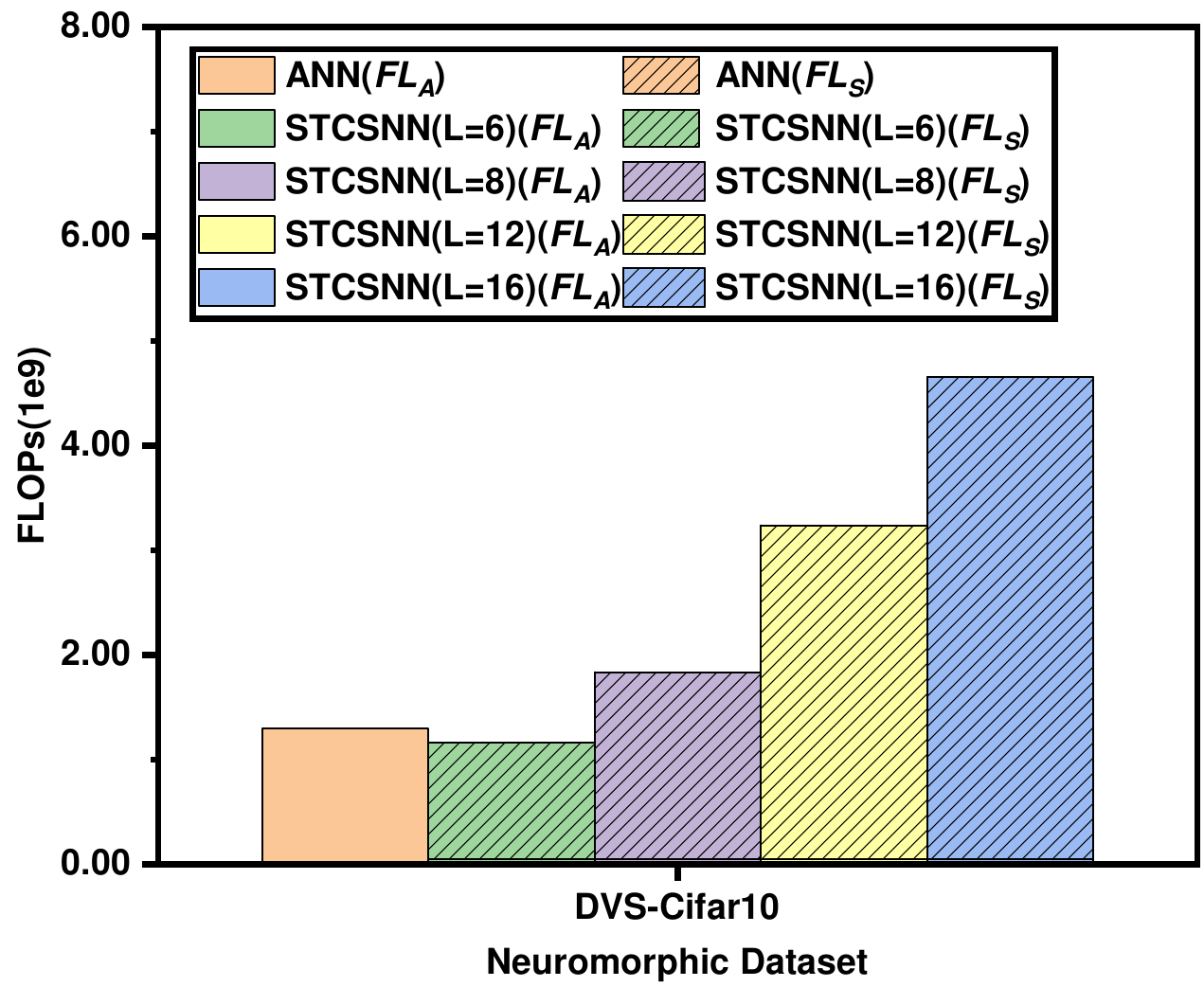}
\label{FLOPsNeuromorphic}
}
\caption{FLOPs of ResNet18 and VGG16. (a) Static Datasets (ResNet18) (b)Neuromorphic dataset (VGG16)}
}
\end{figure*}

\begin{figure}[!t]
\centerline{\includegraphics[width=0.9\columnwidth]{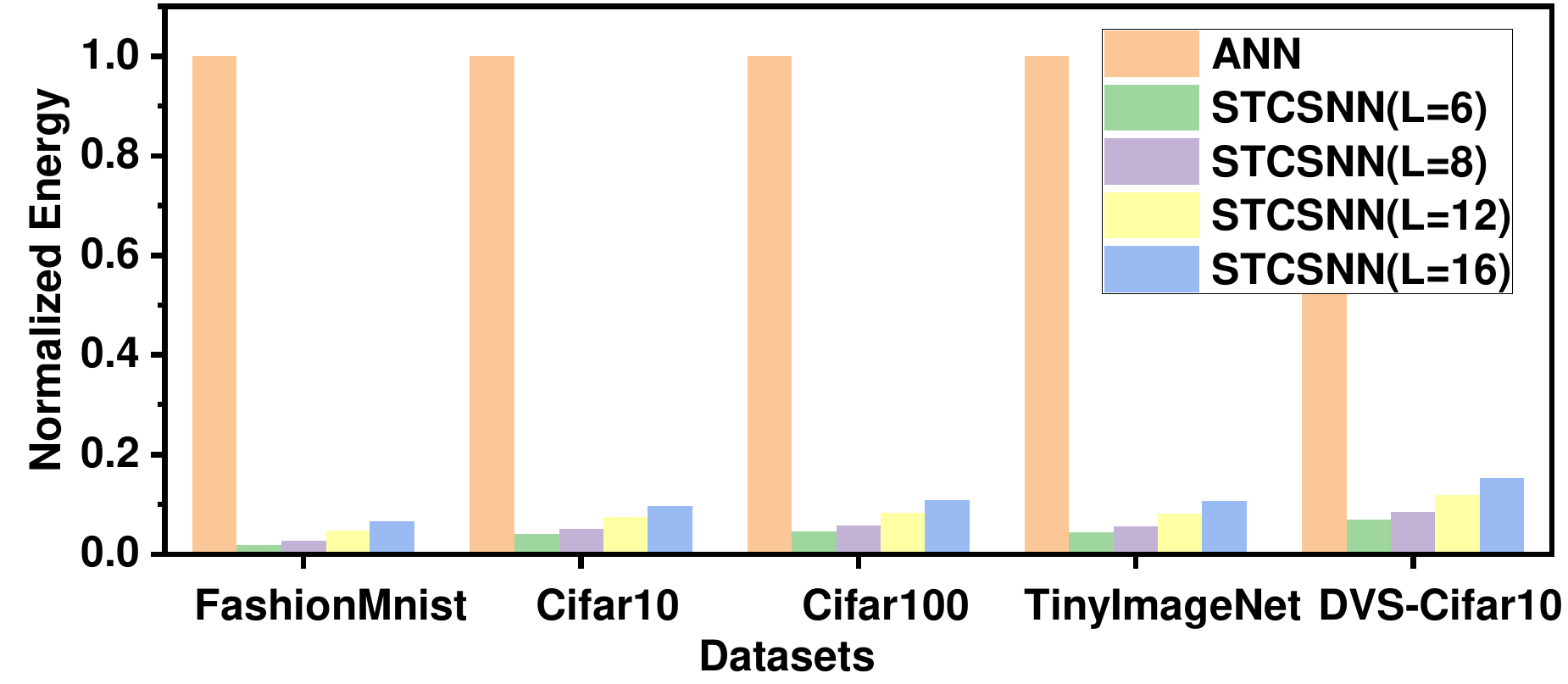}}
\caption{Comparison of computation energy}
\label{Energy}
\end{figure}

\subsubsection{ Floating Point Operations (FLOPs) $\&$ computation Energy}
In this section, we refer \cite{kundu2021towards} to calculate the FLOPs and computation energy of STCSNN.
Consider a convolutional layer l with weight tensor $W^l \in  R^{k^l \times k^l \times C^l_i \times C^l_o}$ takes an input activation tensor $A^l \in R^{k^l \times k^l \times C^l_i}$, where $H^l_i, W^l_i, k^l,$ $C^l_i, C^l_o$ are the input height, width, filter height/width, channel size, and the number of filters, respectively.
$\zeta$ is the average spiking activity (also known as average spike count) of an SNN layer $l$.
The FLOPs of convolutional operations in layer $l$ can be caculated based on Table \ref{T3}.
The FLOPs of other operations, such as batch normalization and non-spiking operations in CAB, are also collected to estimate the power consumption.
\begin{table}[h]
\caption{Convolutional layer FLOPs for ANN and SNN}
\label{T3}
\centering
\resizebox{0.5\textwidth}{!}
{
\begin{tabular}{ll} 
\hline
Model         & FLOPs of a convolutional layer $l$                       \\ 
\hline
ANN       &$(k^l)^2 \times H^l_o \times W^l_o \times C^l_o \times C^l_i$     \\
SNN & $(k^l)^2 \times H^l_o \times W^l_o \times C^l_o \times C^l_i \times \zeta ^l$\\
\hline
\end{tabular}}
\end{table}
The computation energy of ANN and SNNs can be estimated by
\begin{equation}
{E_{ANN}=\sum\nolimits_{l=1}^{L}{FL_{A}^l \cdot {E_{MAC}}}}
\label{eq101}
\end{equation}
\begin{equation}
{E_{SNN}=\sum\nolimits_{l=1}^{L}{FL_{A}^l \cdot {E_{MAC}}} + \sum\nolimits_{l = 2}^L {FL_{S}^l}  \cdot {E_{AC}}}
\label{eq10}
\end{equation}

where $FL_S^l$ is the number of FLOPS in the $l^{th}$ layer, which need AC operations, such as spiking convolutional operations, and $FL_A^l$ is the number of FLOPs in the $l^{th}$ layer, which need MAC operations, such as convolutional operations, CAB block operations, and BN operations, etc.
$E_{MAC}$ and $E_{AC}$ denote the multiply-and-accumulates (MAC) and accumulates (AC) energy, respectively.
To estimate the computation energy, we apply a 45 nm CMOS process at 0.9 V, where $E_{AC}$ = 0.1 pJ, while $E_{MAC}$ = 3.2pJ \cite{R4} for 32-bit integer representation.
For static datasets, we chose ResNet18 as the network architecture.
For the neuromorphic dataset, we apply VGG18 as the network architecture.
Fig. \ref{FLOPsStatic} and \ref{FLOPsNeuromorphic} show the average FLOPs of ResNet18 and VGG16.
Compared with ANN, STCSNNs have fewer FLOPs when the length of the spike train is less than 6.
Though STCSNNs have more FLOPs when the spike train is longer than 6, the accumulation (AC) operation rather than the multiply-and-accumulate (MAC) operation is dominant due to the binary spiking information, which helps STCSNN keep high energy efficiency.
As Fig. \ref{Energy} shows the STCSNNs have significant energy reduction.
The energy efficiency can reach up to $11.76\times \sim 36.79 \times$, when the length of spike train is 8.
Even if the length of the spike train is 16, STCSNN still has $6.54 \times \sim 15.17 \times $ energy reduction.
In Table \ref{T4}, we also compare our method with other SNN methods.
For a fair comparison, we used the same dataset(Cifar10) and architecture(VGG16) as in \cite{kundu2021towards}.
Table \ref{T4} shows that our proposed method achieved 4.8$\%$-5.35$\%$ accuracy improvement with fewer time steps.
Compared with SNN(rate code)\cite{kundu2021towards}, our proposed method has low computation energy due to fewer time steps.
Our proposed method has a higher computation energy than SNN(direct input)\cite{kundu2021towards}.
Compared with SNN(direct input)\cite{kundu2021towards}, Our method has higher computation energy when the time step is 8 because we need to apply non-spiking operations, such as operations in CAB, which have some extra computation energy.
We sacrificed a few energy efficiency characteristics of SNN and achieved significant performance improvements.
When the time step is 6, our proposed STCSNN has lower computation energy but still significantly improves accuracy.
Overall, our proposed STCSNN performs well in performance and power consumption.

\begin{table*}[h]
\caption{Comparison of computation energy with other SNN methods}
\label{T4}
\centering
\resizebox{0.99\textwidth}{!}
{
\begin{tabular}{lllllll} 
\hline
Model & Methods         &Accuracy($\%$) &FLOPs(1E9) &Time Steps &Energy(mJ)  &\makecell{Energy\\Improvement}       \\ 
\hline
ANN   &ANN          &  95.56   &0.13  & 1 & 0.42 & 1$\times$\\ 
Kundu, et al\cite{kundu2021towards}   &SNN(rate code)        &  90.54   & 0.28 & 100 &0.028 &15$\times$\\
Kundu, et al\cite{kundu2021towards}   &SNN(direct input)         &  89.99   & 0.11& 10 &0.011 &38$\times$\\
Our Model   &SNN    &  95.34   & 0.098 & 8 &0.013 &32$\times$\\
Our Model   &SNN    &  93.35   & 0.067 & 6 &0.010 &42$\times$\\
\hline
\end{tabular}
}
\end{table*}

\section{Discussions and Conclusion}
This paper proposes a high energy efficiency spike-train level spiking neural network with spatio temporal conversion. 
Based on the idea of the spatio temporal conversion, spatio-temporal conversion blocks (STCBs) are proposed to keep the low power features of SNNs and improve the accuracy of SNNs.
We chose ResNet18 and VGG16 as network architecture to evaluate the proposed STCSNN on static and neuromorphic datasets, including Fashion-Mnist, Cifar10, Cifar100, TinyImageNet, and DVS-Cifar10.
The experiment results show that our proposed STCSNN outperforms the state-of-the-art accuracy on nearly all datasets, using fewer time steps and being highly energy-efficient.
As for limitations, the proposed STCSNNs may suffer from a certain degree of performance drop when the latency is extremely low(e.g., the length of the spike train is less than 4) since the STCSNNs require enough spikes to transmit information between layers. 
In further work, we will study how to improve the information capacity of spiking trains and neural models to make the SNN obtain a good performance when the length of the spike train is extremely short.
Another area for improvement is that we still need bath normalization in our proposed SNN, which may decrease the advantage in energy efficiency.
We will try to optimize the architecture of our proposed STCSNN to improve its energy efficiency in the future.

\section*{Acknowledgment}
This work was supported by the National Natural Science Foundation of China under Grant 62004146, by the China Postdoctoral Science Foundation funded project under Grant 2021M692498, by the Fundamental Research Funds for the Central Universities under Grant XJSJ23106, and by Science and Technology Projects in Guangzhou under Grant SL2022A04J00095.


\bibliography{mybibfile}

\end{document}